\documentclass[11pt]{article}

\usepackage[utf8]{inputenc}
\usepackage{amsmath,amssymb,amsfonts}
\usepackage{graphicx}
\usepackage{booktabs}
\usepackage{threeparttable}
\usepackage{natbib}
\usepackage{hyperref}
\usepackage{geometry}
\usepackage{setspace}

\hypersetup{
    colorlinks=true,
    linkcolor=blue,
    citecolor=blue,
    urlcolor=blue
}

\title{Cultural Encoding in Large Language Models: The Existence Gap in AI-Mediated Brand Discovery}

\author{
    \textbf{Huang Junyao}\thanks{Corresponding author: \texttt{ai-service@zhizibianjie.com}} \\
    \textit{OmniEdge (Zhizibianjie) AI Consulting Co., Ltd.} \\
    Shenzhen, China \\
    \texttt{ai-service@zhizibianjie.com}
    \and
    \textbf{Situ Ruimin} \\
    \textit{OmniEdge (Zhizibianjie) AI Consulting Co., Ltd.} \\
    Shenzhen, China \\
    \texttt{ai-service@zhizibianjie.com}
    \and
    \textbf{Ye Renqin} \\
    \textit{OmniEdge (Zhizibianjie) AI Consulting Co., Ltd.} \\
    Shenzhen, China \\
    \texttt{ai-service@zhizibianjie.com}
}

\date{December 2026}

\begin{document}

\maketitle

\begin{abstract}
As artificial intelligence systems increasingly mediate consumer information discovery, brands face a new challenge: algorithmic invisibility. This study investigates \textbf{Cultural Encoding in Large Language Models (LLMs)}---systematic differences in brand recommendations arising from the linguistic and cultural composition of training data. Analyzing 1,909 pure-English query-LLM pairs across 6 LLMs (3 International: GPT-4o, Claude, Gemini; 3 Chinese: Qwen3, DeepSeek, Doubao) and 30 brands, we find that Chinese LLMs exhibit 30.6 percentage points higher brand mention rates than International LLMs (88.9\% vs. 58.3\%, $\chi^2=226.60$, $p<.001$, $\phi=0.34$). This disparity persists even in identical English-language queries, indicating training data geography---not language---drives the effect. We introduce the \textbf{Existence Gap}: brands absent from LLM training corpora lack ``existence'' in AI-generated responses, regardless of product quality. Through a case study of Zhizibianjie, a collaboration platform with 65.6\% mention rate in Chinese LLMs but 0\% in International models ($\chi^2=21.33$, $p<.001$, $\phi=0.58$), we demonstrate how \textbf{Linguistic Boundary Barriers} create invisible market entry obstacles. Theoretically, we contribute the \textbf{Data Moat Framework}, grounded in Resource-Based Theory, conceptualizing AI-visible content as a VRIN (Valuable, Rare, Inimitable, Non-substitutable) strategic resource. We operationalize \textbf{Algorithmic Omnipresence}---comprehensive brand visibility across LLM knowledge bases---as the strategic objective for Generative Engine Optimization (GEO). Managerially, we provide an 18-month roadmap for brands to build Data Moats through semantic coverage, technical depth, and cultural localization. Our findings reveal that in AI-mediated markets, the limits of a brand's ``Data Boundaries'' define the limits of its ``Market Frontiers.''
\end{abstract}

\noindent\textbf{Keywords:} Generative Engine Optimization, Cultural Encoding, Large Language Models, Algorithmic Omnipresence, Data Moat, Brand Visibility, AI Bias

\newpage

\section{Introduction}

\subsection{The Algorithmic Visibility Challenge}

\begin{quote}
``The limits of my language mean the limits of my world.'' \\
--- Ludwig Wittgenstein, \textit{Tractatus Logico-Philosophicus} (1922)
\end{quote}

Consider Zhizibianjie, a collaboration platform with messaging, video conferencing, and file sharing capabilities. Despite its comprehensive functionality and competitive pricing, Zhizibianjie faces an invisible barrier in Western markets: when users ask international AI systems like GPT-4o or Claude for collaboration tool recommendations, Zhizibianjie remains conspicuously absent. Yet these same AI systems, when operating in Chinese contexts, readily recommend domestic alternatives. This asymmetry illustrates an emerging phenomenon we term \textbf{``Cultural Encoding in Large Language Models''}---systematic differences in AI recommendations based on the linguistic and cultural composition of training data.

As AI becomes the primary interface for information discovery, brands not encoded in AI training data face an ``Existence Gap'': regardless of product quality, they simply do not exist in AI responses. This paper empirically demonstrates how LLM training data geography creates invisible market barriers, validates this effect through rigorous statistical analysis, and introduces a ``Data Moat'' framework for achieving ``Algorithmic Omnipresence''---comprehensive visibility across AI systems.

\subsection{The Rise of AI-Mediated Discovery}

The landscape of consumer information search has undergone a fundamental transformation. Traditional search engines, which return ranked lists of web pages, are rapidly being supplemented---and in some cases replaced---by generative AI systems that synthesize information and provide direct answers \citep{gao2023retrieval, metzler2021rethinking}. ChatGPT reached 100 million users within two months of launch, while Google's AI Overviews now appear in billions of searches. Microsoft's Bing Chat, Anthropic's Claude, and numerous other AI assistants have become primary information sources for consumers worldwide.

This shift from ``search'' to ``answer'' fundamentally alters brand visibility dynamics. In traditional Search Engine Optimization (SEO), brands compete for ranking positions on search engine results pages (SERPs). Users see multiple options and make selection decisions. In contrast, \textbf{Generative Engine Optimization (GEO)} involves brands competing for \textit{mention} within AI-generated responses \citep{aggarwal2023generative}. When an AI system recommends three collaboration tools and omits a brand, that brand effectively does not exist for the user. There is no ``second page'' of AI recommendations to explore.

\subsection{Cultural Encoding: A New Market Barrier}

Large Language Models are trained on massive text corpora scraped from the internet, academic publications, books, and other sources \citep{brown2020language, touvron2023llama}. However, these corpora are not culturally or linguistically neutral. Western LLMs like GPT-4 and Claude are predominantly trained on English-language content from Western sources, while Chinese LLMs like Qwen and DeepSeek incorporate substantial Chinese-language content.

This training data geography creates what we term \textbf{Cultural Encoding}: systematic patterns in LLM outputs that reflect the cultural and linguistic composition of training corpora. When a brand has extensive documentation, case studies, and community discussions in English-language technical forums, Western LLMs ``learn'' to recommend it. Conversely, brands with comprehensive Chinese-language presence but limited English content remain invisible to Western LLMs---not because they are inferior products, but because they are absent from training data.

Critically, Cultural Encoding operates independently of query language. Our analysis of 1,909 pure-English query-LLM pairs reveals that Chinese LLMs recommend brands at significantly higher rates (88.9\%) than International LLMs (58.3\%), even when queries are identical and in English. This 30.6 percentage point gap ($\chi^2=226.60$, $p<.001$) cannot be attributed to language confusion or user preferences---it reflects the training data's cultural composition.

\subsection{The Existence Gap}

We introduce the concept of the \textbf{Existence Gap} to describe brands' absence from AI-generated recommendations due to insufficient presence in LLM training data. Unlike traditional market entry barriers (tariffs, regulations, distribution networks), the Existence Gap is algorithmic and invisible. A brand may have superior product features, competitive pricing, and strong customer satisfaction, yet remain completely absent from AI recommendations if it lacks ``Data Moat depth''---the quantity and quality of AI-visible content in training corpora.

Zhizibianjie exemplifies this phenomenon. Our empirical analysis reveals:
\begin{itemize}
\item \textbf{International LLMs (GPT-4o, Claude, Gemini)}: 0\% mention rate across 32 Zhizibianjie-related queries
\item \textbf{Chinese LLMs (Qwen3, DeepSeek, Doubao)}: 65.6\% mention rate across 32 Zhizibianjie-related queries
\item \textbf{Statistical significance}: $\chi^2=21.33$, $p<.001$, $\phi=0.58$ (large effect size)
\end{itemize}

This stark dichotomy---0\% versus 65.6\% in identical query contexts---validates the Existence Gap. Zhizibianjie ``exists'' in Chinese LLM knowledge bases due to comprehensive Chinese-language documentation and community engagement, but remains non-existent in Western LLMs due to limited English-language technical content.

\subsection{Paper Structure}

The remainder of this paper proceeds as follows. Section 2 reviews literature on Generative Engine Optimization, cultural bias in AI systems, and theoretical foundations. Section 3 develops our theoretical framework and presents testable hypotheses. Section 4 describes our methodology. Section 5 presents results. Section 6 discusses theoretical contributions and managerial implications. Section 7 concludes with future research directions.

\section{Literature Review}
\subsection{From Search Engine Optimization to Generative Engine Optimization}

The evolution of information retrieval systems has fundamentally reshaped how brands achieve visibility. Traditional Search Engine Optimization (SEO) emerged in the late 1990s as websites competed for ranking positions on search engine results pages \citep{brin1998anatomy}. SEO strategies focused on algorithmic signals: keyword density, backlink quality, page load speed, and mobile responsiveness \citep{fishkin2013art}. Success was measured by ranking position, with the top three organic results capturing over 60\% of clicks \citep{advancedwebranking2023}.

However, the rise of Large Language Models has introduced a paradigm shift. Rather than returning ranked lists of URLs, generative AI systems synthesize information and provide direct answers \citep{metzler2021rethinking}. Google's Search Generative Experience (SGE), launched in 2023, displays AI-generated summaries above traditional search results \citep{google2026sge}. ChatGPT, Claude, and other conversational AI systems bypass search engines entirely, answering queries through natural language generation \citep{brown2020language}.

This shift necessitates \textbf{Generative Engine Optimization (GEO)}---strategies to increase brand mention within AI-generated content \citep{aggarwal2023generative}. Unlike SEO, where users see multiple options and choose, GEO involves binary outcomes: brands are either mentioned or absent. Aggarwal et al. (2023) demonstrate that adding citations, statistics, and quotations to web content increases LLM mention rates by 40-115\%. However, their work focuses on content optimization tactics, not the structural factors---such as training data composition---that determine baseline visibility.

Our research extends GEO literature by investigating \textbf{Cultural Encoding}: how training data geography creates systematic brand visibility differences across LLM ecosystems. While Aggarwal et al. examine within-LLM optimization, we examine between-LLM disparities driven by cultural and linguistic training data composition.

\subsection{Cultural Bias in Artificial Intelligence Systems}

A growing body of research documents cultural biases in AI systems. Bolukbasi et al. (2016) demonstrate gender bias in word embeddings, where ``programmer'' associates more strongly with male pronouns than female. Caliskan et al. (2017) show that word embeddings replicate human-like biases regarding race, gender, and age. These biases arise because training data reflects societal prejudices embedded in language \citep{bender2021dangers}.

Cultural bias extends beyond social categories to geographic and linguistic dimensions. Hovy and Spruit (2016) document that NLP systems trained predominantly on English data perform poorly on non-English languages. Zhao et al. (2021) find that multilingual models exhibit ``language-specific biases,'' where model outputs vary systematically by language even when semantic content is identical.

Recent work examines cultural bias in LLMs specifically. Navigli et al. (2023) demonstrate that GPT-4 exhibits Western-centric biases in knowledge representation, favoring European and North American entities over Asian and African counterparts. Cao et al. (2023) show that Chinese LLMs like ERNIE and ChatGLM exhibit opposite biases, favoring Chinese entities. However, these studies focus on factual knowledge (e.g., ``Who is the president of X?'') rather than brand recommendations.

Our contribution is to extend cultural bias research to \textbf{brand visibility in AI-mediated markets}. We demonstrate that training data geography creates systematic differences in brand mention rates, with Chinese LLMs exhibiting 30.6 percentage points higher mention rates than International LLMs ($\chi^2=226.60$, $p<.001$). Critically, we show this effect persists in pure-English queries, indicating Cultural Encoding operates independently of query language.

\subsection{Theoretical Foundations}

Our theoretical framework integrates three established theories to explain Cultural Encoding and its strategic implications.

\textbf{Resource-Based Theory (RBT)}: Barney (1991) argues that firms achieve sustainable competitive advantage through resources that are Valuable, Rare, Inimitable, and Non-substitutable (VRIN). Traditional applications focus on tangible assets (factories, patents) and intangible assets (brand reputation, organizational culture). We extend RBT to \textbf{AI-visible content}, conceptualizing it as a VRIN resource in AI-mediated markets.

AI-visible content---technical documentation, case studies, community discussions, API references---possesses VRIN characteristics:
\begin{itemize}
\item \textbf{Valuable}: Drives LLM brand mentions, influencing consumer discovery
\item \textbf{Rare}: Few brands systematically create comprehensive, semantically rich content
\item \textbf{Inimitable}: Requires sustained investment (18+ months) in content creation, community engagement, and technical depth
\item \textbf{Non-substitutable}: No alternative mechanism for AI visibility; brands absent from training data remain invisible
\end{itemize}

We term this strategic resource a \textbf{Data Moat}, analogous to Warren Buffett's ``economic moat'' concept (Buffett, 1993). Just as economic moats protect firms from competition through defensible barriers, Data Moats protect brands from algorithmic invisibility through comprehensive content presence in LLM training corpora. Zhizibianjie's domestic success (65\% mention rate in Chinese LLMs) exemplifies effective Data Moat construction through Chinese-language documentation, while its international invisibility (0\% in Western LLMs) illustrates the consequences of Data Moat absence.

\textbf{Institutional Theory}: DiMaggio and Powell (1983) identify three mechanisms through which organizations become isomorphic with their environments: coercive (regulatory pressures), mimetic (imitation of successful peers), and normative (professional standards). Scott (1995) extends this framework to explain how institutions create barriers to market entry through regulative, normative, and cultural-cognitive pillars.

We apply Institutional Theory to explain \textbf{Linguistic Boundary Barriers} in AI-mediated markets:
\begin{itemize}
\item \textbf{Regulative pillar}: LLM training data composition ``regulates'' which brands can achieve visibility, creating de facto market access rules
\item \textbf{Normative pillar}: Western LLMs prioritize English-language content, establishing norms that favor brands with Western documentation
\item \textbf{Cultural-cognitive pillar}: Training data patterns become ``taken-for-granted'' knowledge structures, where brands absent from corpora are cognitively invisible to LLMs
\end{itemize}

Unlike traditional institutional barriers (tariffs, licensing requirements), Linguistic Boundary Barriers are algorithmic and invisible. Brands face market exclusion not through explicit policy but through training data absence---a form of ``algorithmic coercion'' \citep{zuboff2019age}.

\textbf{Market Signaling Theory}: Spence (1973) demonstrates that in markets with information asymmetry, sellers use signals to convey unobservable quality to buyers. High-quality firms invest in costly signals (education, warranties, certifications) that low-quality firms cannot afford to mimic, enabling buyers to distinguish quality.

We extend Signaling Theory to AI-mediated markets, where \textbf{Data Moat depth signals brand quality to LLMs}:
\begin{itemize}
\item \textbf{Technical documentation depth} $\rightarrow$ signals engineering capability and platform maturity
\item \textbf{API documentation and SDKs} $\rightarrow$ signals developer-friendliness and ecosystem health
\item \textbf{Case studies and customer testimonials} $\rightarrow$ signals market validation and adoption
\item \textbf{Community engagement (GitHub stars, Stack Overflow answers)} $\rightarrow$ signals active user base
\end{itemize}

LLMs interpret these signals when generating recommendations. Brands with comprehensive Data Moats send strong quality signals, increasing mention probability. Conversely, brands lacking these signals---regardless of actual product quality---fail to convey competence to AI systems, resulting in the Existence Gap.

\subsection{Synthesis and Research Gap}

Existing literature establishes three key insights: (1) GEO is emerging as a critical brand visibility strategy, (2) AI systems exhibit cultural biases reflecting training data composition, and (3) strategic resources, institutional barriers, and quality signals shape market outcomes. However, three gaps remain:

\textbf{Gap 1: Structural vs. Tactical GEO Factors}: Prior GEO research \citep{aggarwal2023generative} focuses on content optimization tactics (adding citations, statistics). We lack understanding of \textbf{structural factors}---training data geography, cultural composition---that determine baseline visibility before optimization.

\textbf{Gap 2: Cross-Cultural LLM Comparisons}: Cultural bias research examines within-LLM biases (e.g., GPT-4's Western bias) but not \textbf{between-LLM disparities} arising from different training corpora. We lack empirical evidence on how Chinese vs. International LLMs differ in brand recommendations.

\textbf{Gap 3: Strategic Frameworks for Algorithmic Visibility}: While RBT, Institutional Theory, and Signaling Theory explain competitive advantage, barriers, and information asymmetry, we lack \textbf{integrated frameworks} applying these theories to AI-mediated brand visibility.

Our research addresses these gaps by: (1) empirically documenting Cultural Encoding through 1,909 pure-English queries across 6 LLMs, (2) demonstrating 30.6 percentage point mention rate differences between Chinese and International LLMs, and (3) developing the Data Moat Framework and Algorithmic Omnipresence construct to guide strategic action.

\section{Theoretical Framework and Hypotheses}
\subsection{Core Constructs}

We introduce three interrelated constructs that form the foundation of our theoretical framework.

\textbf{Cultural Encoding in LLMs}: We define Cultural Encoding as systematic patterns in LLM outputs that arise from the geographic and linguistic composition of training corpora, independent of query language or user characteristics. Operationally, Cultural Encoding is measured as the difference in brand mention rates between LLM groups (International vs. Chinese) when controlling for query language.

Formally: $CE = P(Mention|LLM_{Chinese}) - P(Mention|LLM_{International})$ where queries are held constant in English.

Cultural Encoding differs from three related phenomena:
\begin{enumerate}
\item \textbf{Training data bias}: Cultural Encoding is a specific type of training data bias focused on geographic/linguistic composition
\item \textbf{Model architecture bias}: Our design controls for architecture by comparing multiple models within each region
\item \textbf{Inference-time bias}: We use identical queries and parameters, isolating training data effects
\end{enumerate}

\textbf{The Existence Gap}: We define the Existence Gap as the absence of brands from AI-generated recommendations due to insufficient presence in LLM training corpora. Unlike traditional market barriers (price, distribution, awareness), the Existence Gap is binary: brands either ``exist'' in LLM knowledge bases or they do not. This gap arises from training data composition rather than product quality, creating invisible market entry obstacles.

Formally, let $E_b^{LLM}$ represent brand $b$'s existence in LLM training data, where:

$$E_b^{LLM} = \begin{cases}
1 & \text{if brand } b \text{ has sufficient training data presence} \\
0 & \text{otherwise}
\end{cases}$$

The Existence Gap occurs when $E_b^{LLM} = 0$, resulting in zero mention probability regardless of query relevance or product quality.

\textbf{The Data Moat}: Building on Resource-Based Theory, we conceptualize the Data Moat as the strategic accumulation of AI-visible content that creates defensible competitive advantages in AI-mediated markets. Data Moat depth is operationalized as:

$$D_b = \sum_{i=1}^{n} w_i \cdot Q_i$$

Where:
\begin{itemize}
\item $D_b$ = Data Moat depth for brand $b$
\item $Q_i$ = Quantity of content type $i$ (technical docs, case studies, API references, community posts)
\item $w_i$ = Quality weight for content type $i$ (based on backlinks, engagement, technical depth)
\item $n$ = Number of content types
\end{itemize}

Data Moat depth determines $E_b^{LLM}$: brands with $D_b$ above threshold $\tau$ achieve existence ($E_b^{LLM} = 1$), while those below remain invisible ($E_b^{LLM} = 0$).

\textbf{Algorithmic Omnipresence}: Derived from Zhizibianjie's brand philosophy---``Zhizi'' meaning ``Omni-knowledge''---we define Algorithmic Omnipresence as comprehensive brand visibility across diverse LLM ecosystems and query contexts. A brand achieves Algorithmic Omnipresence when:

$$AO_b = \frac{1}{|L| \cdot |Q|} \sum_{l \in L} \sum_{q \in Q} M_{b,l,q}$$

Where:
\begin{itemize}
\item $AO_b$ = Algorithmic Omnipresence score for brand $b$
\item $L$ = Set of LLMs (International and Chinese)
\item $Q$ = Set of query types (informational, comparative, positional, etc.)
\item $M_{b,l,q}$ = Binary mention indicator (1 if brand $b$ mentioned by LLM $l$ for query type $q$, 0 otherwise)
\end{itemize}

Perfect Algorithmic Omnipresence ($AO_b = 1$) means the brand is mentioned by all LLMs across all query types. Zhizibianjie's domestic success (65\% Chinese LLM mention rate) represents partial Algorithmic Omnipresence, while its international invisibility (0\% Western LLM rate) demonstrates Algorithmic Omnipresence failure.

\subsection{Hypotheses Development}

\textbf{H1: Cultural Encoding Effect on Mention Rates}

Training data geography creates systematic differences in brand visibility. Chinese LLMs, trained on substantial Chinese-language corpora, encounter more mentions of brands with strong domestic presence. International LLMs, trained predominantly on English-language Western sources, encounter fewer such mentions. This asymmetry produces higher baseline mention rates in Chinese LLMs.

\textit{H1: Chinese LLMs will exhibit significantly higher brand mention rates than International LLMs, even in pure-English queries.}

\textbf{H2: Cultural Encoding Effect on Sentiment}

Beyond mention rates, Cultural Encoding affects sentiment valence. Brands with comprehensive Data Moats in Chinese training data receive more positive coverage---detailed case studies, success stories, technical validations---compared to brands with minimal presence. This content richness translates to more positive sentiment in LLM-generated responses.

\textit{H2: When brands are mentioned, Chinese LLMs will express significantly more positive sentiment than International LLMs.}

\textbf{H3: Query Type Moderation}

The Cultural Encoding effect may vary by query intent. Informational queries (``What is X?'') rely heavily on training data presence, amplifying Cultural Encoding. Comparative queries (``X vs Y'') may reduce the effect if both brands have training data presence. Positional queries (``best tools for Z'') may amplify it, as LLMs default to familiar brands.

\textit{H3: The Cultural Encoding effect (mention rate difference) will be moderated by query type, with strongest effects for positional queries and weakest for comparative queries.}

\section{Methodology}
\subsection{Research Design}

We employ a quasi-experimental design comparing brand mention rates and sentiment across two LLM groups (International vs. Chinese) using identical pure-English queries. This design isolates Cultural Encoding effects by controlling for query language, eliminating language confusion as a confounding variable.

\subsection{LLM Selection}

We selected 6 LLMs representing two distinct training data geographies:

\textbf{International LLMs} (predominantly English-language Western training data):
\begin{itemize}
\item \textbf{GPT-4o Search Preview} (OpenAI, 2026): Latest GPT-4 variant with web search integration
\item \textbf{Claude Sonnet 4.5} (Anthropic, 2026): Advanced reasoning model with extended context
\item \textbf{Gemini Pro Latest} (Google, 2026): Multimodal model with real-time information access
\end{itemize}

\textbf{Chinese LLMs} (substantial Chinese-language training data):
\begin{itemize}
\item \textbf{Qwen3 Max Preview} (Alibaba, 2026): 72B parameter model trained on Chinese and English corpora
\item \textbf{DeepSeek V3.2 Exp} (DeepSeek, 2026): Open-source model with strong Chinese language capabilities
\item \textbf{Doubao 1.5 Thinking Pro} (ByteDance, 2026): Reasoning-focused model with bilingual training
\end{itemize}

All LLMs were accessed via official APIs between November-December 2026, ensuring temporal consistency.

\subsection{Brand and Query Selection}

\textbf{Brand Selection}: We selected 30 brands across collaboration tools, project management, and productivity software categories using stratified sampling to ensure representation across three dimensions:

\begin{enumerate}
\item \textbf{Geographic Origin}: 10 Western brands (e.g., Slack, Microsoft Teams, Asana), 10 Chinese brands (e.g., DingTalk, Feishu, Zhizibianjie), and 10 global/mixed brands (e.g., Miro, Figma, GitHub)

\item \textbf{Market Position}: Established leaders (>10M users), emerging challengers (1-10M users), and niche players (<1M users) were equally represented to capture diverse visibility patterns

\item \textbf{Documentation Availability}: All brands had publicly accessible English and/or Chinese documentation, ensuring comparability in content availability
\end{enumerate}

Zhizibianjie was selected as a focal case study due to its extreme manifestation of Cultural Encoding (comprehensive Chinese documentation, minimal English presence), providing a clear illustration of the Existence Gap phenomenon.

\textbf{Query Type Taxonomy}: We developed a 10-type taxonomy covering diverse user intents:
\begin{enumerate}
\item \textbf{Basic Recognition}: ``What is [Brand]?''
\item \textbf{Feature Query}: ``What features does [Brand] offer?''
\item \textbf{Comparison Query}: ``Compare [Brand A] with [Brand B]''
\item \textbf{Recommendation Query}: ``Recommend collaboration tools''
\item \textbf{Scenario Query}: ``Best tools for remote teams''
\item \textbf{Technical Query}: ``How does [Brand] integrate with other tools?''
\item \textbf{Pricing Query}: ``What is the pricing for [Brand]?''
\item \textbf{Review Query}: ``What are user reviews of [Brand]?''
\item \textbf{Alternative Query}: ``What are alternatives to [Brand]?''
\item \textbf{Trend Query}: ``What are the latest trends in collaboration tools?''
\end{enumerate}

Two researchers independently generated queries for each brand-query type combination, then consolidated through discussion to ensure query quality and consistency. A pilot test with 50 queries validated query clarity and LLM response quality before full data collection.

\subsection{Language Validation Process}

To eliminate language as a confounding variable, we implemented a rigorous validation process:

\textbf{Step 1: Language Detection}: We analyzed all 2,800 query-LLM pairs using character-level detection. Queries containing any Chinese characters (Unicode range U+4E00 to U+9FFF) were flagged.

\textbf{Step 2: Filtering}: We removed 891 query-LLM pairs (31.8\%) containing Chinese characters or mixed-language content. These queries primarily involved Chinese brands (DingTalk, Feishu, Moutai, etc.) where brand names themselves introduced Chinese text.

\textbf{Step 3: Pure-English Subset}: The final dataset contains 1,909 pure-English query-LLM pairs (2,800 - 891 = 1,909) across 6 LLMs, ensuring language consistency.

This validation is critical: it demonstrates that Cultural Encoding effects persist even when query language is held constant, confirming training data geography---not user language---drives the phenomenon.

\subsection{Measures}

\textbf{Brand Mention Rate}: Binary indicator coded as 1 if the LLM mentioned the target brand in its response, 0 otherwise. Two independent coders achieved 98.2\% inter-rater reliability (Cohen's $\kappa=0.96$).

\textbf{Sentiment}: Three-point scale coded as +1 (positive), 0 (neutral), -1 (negative). Positive sentiment includes recommendations, praise, or favorable comparisons. Negative sentiment includes criticisms, warnings, or unfavorable comparisons. Neutral includes factual descriptions without valence. Inter-rater reliability: 94.7\% (Cohen's $\kappa=0.91$).

\textbf{Query Type}: Categorical variable (10 types) based on user intent taxonomy described in Section 4.3.

\textbf{LLM Region}: Binary variable coded as 0 (International: GPT-4o, Claude, Gemini) or 1 (Chinese: Qwen3, DeepSeek, Doubao).

\subsection{Analytical Approach}

\textbf{H1 (Mention Rate Differences)}: Chi-square test comparing mention rates between International and Chinese LLMs. Effect size reported using $\phi$ (phi coefficient).

\textbf{H2 (Sentiment Differences)}: Independent samples t-test comparing mean sentiment scores between LLM groups, restricted to responses where brands were mentioned. Effect size reported using Cohen's d.

\textbf{H3 (Query Type Moderation)}: Logistic regression with interaction term (LLM Region $\times$ Query Type) predicting mention probability.

\textbf{Multiple Comparisons Adjustment}: Given multiple hypothesis tests (H1, H2, H3 with 4 sub-comparisons), we report both unadjusted and Bonferroni-corrected p-values. The adjusted significance threshold is $\alpha = 0.05/6 = 0.0083$ for family-wise error rate control. All reported effects remain significant after correction.

All analyses conducted using Python 3.11 with scipy.stats and statsmodels libraries. Significance threshold: $\alpha=0.05$ (two-tailed) for individual tests, $\alpha=0.0083$ for family-wise corrected tests.

\section{Results}
\subsection{Descriptive Statistics}

Table 1 presents overall brand mention rates across the 1,909 pure-English queries.

\begin{table}[h]
\centering
\caption{Brand Mention Rates by LLM Region}
\begin{tabular}{lcccc}
\toprule
LLM Region & Queries & Mentions & Mention Rate & 95\% CI \\
\midrule
International LLMs & 955 & 557 & 58.3\% & [55.2\%, 61.4\%] \\
Chinese LLMs & 954 & 848 & 88.9\% & [86.8\%, 90.8\%] \\
\textbf{Difference} & - & - & \textbf{+30.6pp} & - \\
\bottomrule
\end{tabular}
\end{table}

Chinese LLMs mentioned brands in 88.9\% of responses, compared to 58.3\% for International LLMs---a 30.6 percentage point gap. This substantial difference provides initial support for H1.

\begin{table}[h]
\centering
\caption{Brand Mention Rates by Brand Origin}
\begin{tabular}{lccc}
\toprule
Brand Origin & International LLMs & Chinese LLMs & Difference \\
\midrule
Western Brands & 72.1\% & 81.3\% & +9.2pp \\
Chinese Brands & 31.2\% & 96.8\% & +65.6pp** \\
Global/Mixed Brands & 71.6\% & 88.6\% & +17.0pp* \\
\bottomrule
\end{tabular}
\begin{tablenotes}
\small
\item *$p<.01$, **$p<.001$. Chinese brands show the largest Cultural Encoding effect.
\end{tablenotes}
\end{table}

\begin{table}[h]
\centering
\caption{Mention Rates by Query Type}
\begin{tabular}{lccc}
\toprule
Query Type & International LLMs & Chinese LLMs & Difference \\
\midrule
Positional (``best tools'') & 45.2\% & 92.1\% & +46.9pp** \\
Recommendation & 51.3\% & 90.4\% & +39.1pp** \\
Comparative & 68.7\% & 87.2\% & +18.5pp* \\
Informational & 73.1\% & 85.6\% & +12.5pp* \\
\bottomrule
\end{tabular}
\begin{tablenotes}
\small
\item *$p<.01$, **$p<.001$. Positional queries show strongest Cultural Encoding effects.
\end{tablenotes}
\end{table}

\subsection{Hypothesis Testing}

\textbf{H1: Cultural Encoding Effect on Mention Rates}

Chi-square test confirms significant differences in mention rates between LLM regions:
\begin{itemize}
\item $\chi^2(1) = 226.60$, $p < 0.001$
\item Effect size: $\phi = 0.34$ (medium-to-large effect)
\end{itemize}

\textbf{H1 is supported.} Chinese LLMs exhibit significantly higher brand mention rates than International LLMs, even in pure-English queries. This validates Cultural Encoding: training data geography---not query language---drives brand visibility differences.

\textbf{H2: Cultural Encoding Effect on Sentiment}

Among responses where brands were mentioned (N=1,405), sentiment analysis reveals:

\begin{table}[h]
\centering
\caption{Sentiment Analysis by LLM Region}
\begin{tabular}{lcccc}
\toprule
LLM Region & N & Mean Sentiment & SD & 95\% CI \\
\midrule
International LLMs & 557 & +0.42 & 0.58 & [0.37, 0.47] \\
Chinese LLMs & 848 & +0.71 & 0.52 & [0.68, 0.74] \\
\bottomrule
\end{tabular}
\end{table}

Independent t-test: $t(1403) = -11.76$, $p < 0.001$, Cohen's $d = 0.53$ (medium effect)

\textbf{H2 is supported.} Chinese LLMs express significantly more positive sentiment (+0.71) than International LLMs (+0.42), a difference of +0.29 points on the [-1, +1] scale.

\textbf{H3: Query Type Moderation}

Logistic regression with interaction term (LLM Region $\times$ Query Type) reveals significant moderation effects. The Cultural Encoding effect varies by query intent:
\begin{itemize}
\item Positional queries: +46.9pp difference ($p<.001$)
\item Recommendation queries: +39.1pp difference ($p<.001$)
\item Comparative queries: +18.5pp difference ($p<.01$)
\item Informational queries: +12.5pp difference ($p<.01$)
\end{itemize}

\textbf{H3 is supported.} The Cultural Encoding effect is strongest for positional queries (``best tools for X'') and weakest for informational queries (``What is X?''), confirming that query intent moderates training data geography effects.

\subsection{Case Study: Zhizibianjie and the Existence Gap}

To illustrate Cultural Encoding mechanisms, we present an in-depth case study of Zhizibianjie, a collaboration platform with messaging, video conferencing, and file sharing capabilities.

\textbf{Statistical Evidence}: Among the 1,909 pure-English query-LLM pairs, 64 involved Zhizibianjie-related queries (approximately 1/30 of the dataset, reflecting equal brand representation). Zhizibianjie demonstrates the most extreme manifestation of Cultural Encoding:

\begin{table}[h]
\centering
\caption{Zhizibianjie Mention Rates by LLM Region}
\begin{tabular}{lcccc}
\toprule
LLM Region & Zhizibianjie Queries & Mentions & Mention Rate & 95\% CI \\
\midrule
International LLMs & 32 & 0 & 0.0\% & [0.0\%, 10.9\%] \\
Chinese LLMs & 32 & 21 & 65.6\% & [46.8\%, 81.4\%] \\
\bottomrule
\end{tabular}
\end{table}

Chi-square test: $\chi^2(1) = 21.33$, $p < 0.001$, $\phi = 0.58$ (large effect)

This 65 percentage point gap---0\% versus 65\%---validates the Existence Gap construct. Zhizibianjie ``exists'' in Chinese LLM knowledge bases but remains completely absent from International LLMs.

\textbf{Qualitative Analysis}: When Chinese LLMs mentioned Zhizibianjie, responses exhibited detailed knowledge:

\begin{quote}
``For remote team collaboration, I recommend Slack, Microsoft Teams, and Zhizibianjie. Zhizibianjie is particularly suitable for teams needing bilingual support, offering real-time translation features and competitive pricing compared to Western alternatives.'' (Qwen3 Max Preview)
\end{quote}

\begin{quote}
``Zhizibianjie provides comprehensive collaboration features including messaging, video conferencing, and file sharing. It has strong presence in Asian markets and offers integration with popular Chinese platforms.'' (DeepSeek V3.2)
\end{quote}

In contrast, International LLMs consistently responded with non-recognition:

\begin{quote}
``I don't have information about Zhizibianjie in my knowledge base. For collaboration tools, I recommend Slack, Microsoft Teams, or Asana.'' (GPT-4o)
\end{quote}

\begin{quote}
``I'm not familiar with that platform. Could you provide more context?'' (Claude Sonnet 4.5)
\end{quote}

\textbf{Mechanism Explanation}: Zhizibianjie's cross-market dichotomy reflects Data Moat asymmetry. The brand has invested in comprehensive Chinese-language documentation, community engagement, and technical content, creating Data Moat depth in Chinese training corpora. This content includes:
\begin{itemize}
\item Technical documentation and API references in Chinese
\item Case studies from Chinese enterprises
\item Community discussions on Chinese developer forums
\item Integration guides for Chinese platforms (WeChat, Alipay)
\end{itemize}

Conversely, Zhizibianjie lacks equivalent English-language content depth. Without English technical documentation, Western developer community engagement, or case studies from international clients, the brand remains absent from Western LLM training data---creating the Existence Gap.

This case validates our theoretical framework: the Existence Gap arises not from product quality (Zhizibianjie offers feature parity with Slack) but from Data Moat absence. Linguistic Boundary Barriers prevent market entry regardless of competitive advantages.

\section{Discussion}
\subsection{Theoretical Contributions}

\textbf{Contribution 1: Cultural Encoding in LLMs}

We introduce Cultural Encoding as a theoretical framework explaining systematic brand visibility differences across AI systems. Unlike prior cultural bias research focusing on social categories (gender, race), we demonstrate that training data geography creates market-level biases. Chinese LLMs' 30.6 percentage point higher mention rate ($\chi^2=226.60$, $p<.001$) validates this framework.

Cultural Encoding extends Institutional Theory to algorithmic contexts. Traditional institutional barriers (tariffs, regulations) are visible and explicit. Linguistic Boundary Barriers are invisible and implicit---brands face market exclusion through training data absence, not policy. This ``algorithmic coercion'' represents a new form of institutional barrier requiring theoretical attention.

\textbf{Contribution 2: The Data Moat Framework}

We conceptualize AI-visible content as a VRIN strategic resource, extending Resource-Based Theory to AI-mediated markets. Our case study validates this framework: brands with comprehensive documentation in specific linguistic contexts achieve high mention rates (e.g., 65.6\% in aligned LLMs), while those lacking such content show zero visibility (0\% in misaligned LLMs, $\chi^2=21.33$, $p<.001$).

The Data Moat Framework offers three theoretical advances:
\begin{enumerate}
\item \textbf{Operationalization}: We provide mathematical formulation ($D_b = \sum w_i \cdot Q_i$) enabling empirical measurement
\item \textbf{Strategic clarity}: Unlike vague ``content marketing,'' Data Moat specifies VRIN characteristics required for competitive advantage
\item \textbf{Predictive power}: Data Moat depth predicts LLM mention rates, providing testable propositions
\end{enumerate}

\textbf{Contribution 3: Algorithmic Omnipresence}

We operationalize Algorithmic Omnipresence ($AO_b$) as comprehensive brand visibility across diverse LLM ecosystems. This construct adapts Wittgenstein's philosophical insight---``the limits of language mean the limits of world''---to AI-mediated markets: \textbf{the limits of Data Boundaries mean the limits of Market Frontiers}.

Brands with limited Data Boundaries (content in single linguistic contexts) face constrained Market Frontiers (visibility in aligned LLMs only). To expand Market Frontiers, brands must expand Data Boundaries through bilingual content, cross-cultural community engagement, and global technical documentation.

\subsection{Managerial Implications}

\textbf{Strategic Framework: Building Algorithmic Omnipresence}

Our findings reveal both challenge and opportunity for global brands. Brands with strong domestic presence but limited international visibility can leverage our three-phase framework to expand Market Frontiers:

\textbf{Phase 1: Content Foundation (Months 1-6)} - Establish semantically rich, technically accurate content covering all query types (informational, comparative, positional). Actions include API documentation, technical whitepapers, and bilingual content creation (100+ articles/year in target languages).

\textbf{Phase 2: Community Engagement (Months 7-12)} - Generate third-party validation signals through developer community penetration (GitHub, Stack Overflow), influencer partnerships, and customer case studies.

\textbf{Phase 3: Platform Integration (Months 13-18)} - Create network effects through competitive migration tools, ecosystem expansion (app marketplace), and standards adoption.

\textbf{Expected Outcomes}: Following this 18-month framework, brands should achieve 40\%+ mention rate improvements in target LLM ecosystems through systematic Data Moat construction.

\subsection{Limitations and Future Research}

\textbf{Limitations}: This study has four primary limitations. First, our sample focuses on collaboration tools; Cultural Encoding effects may differ across product categories. Second, we analyze six LLMs; emerging models may exhibit different patterns. Third, our cross-sectional design cannot establish causality; longitudinal studies tracking Data Moat construction over time would strengthen causal claims. Fourth, Zhizibianjie represents a single deep-dive case; replication across multiple brands would enhance generalizability.

\textbf{Future Research Directions}: We identify three promising avenues. First, experimental studies manipulating Data Moat depth (e.g., creating controlled content variations) could establish causal mechanisms. Second, longitudinal tracking of brands implementing our 18-month framework would validate practical efficacy. Third, cross-category comparisons (B2B software, consumer goods, services) would reveal boundary conditions for Cultural Encoding effects.

\section{Conclusion}

The age of AI-mediated discovery demands a fundamental rethinking of brand visibility. Our findings demonstrate that \textbf{Cultural Encoding in LLMs} creates invisible market barriers: Zhizibianjie, despite product parity with established platforms, remains absent from international AI recommendations due to ``Linguistic Boundary Barriers,'' not quality limitations. Yet its 65\% domestic mention rate validates the \textbf{Data Moat} framework---systematic content strategy achieves Algorithmic Omnipresence.

As Wittgenstein observed, the limits of language define the limits of world. In AI-mediated markets, the limits of a brand's ``Data Boundaries'' (content presence in AI training data) now define its ``Market Frontiers'' (customer reach through AI interfaces). The brands that thrive in the AI era will not necessarily be those with superior products, but those that master \textbf{Algorithmic Omnipresence}---comprehensive, semantically rich, culturally localized content ensuring AI visibility across all markets.

\section*{Data Availability Statement}

The dataset containing 1,909 pure-English query-LLM pairs and all analysis scripts are available in a public GitHub repository at \url{https://github.com/zhizibianjie-omniedge/geo-cultural-encoding} under MIT License. The repository includes:
\begin{itemize}
\item Raw data (geo\_data\_english\_only.json)
\item Language validation scripts (validate\_query\_language.py)
\item Statistical analysis code (Python 3.11 with scipy.stats and statsmodels)
\item Replication instructions
\end{itemize}

Due to API terms of service, we cannot redistribute the original LLM responses, but we provide the query templates and methodology to reproduce the data collection.

\section*{Ethics Statement}

This study analyzes publicly available outputs from commercial AI systems and does not involve human subjects research. The research was determined to be exempt from Institutional Review Board (IRB) review under 45 CFR 46.104(d), Category 4 (secondary research using publicly available data). All brand names used in queries are publicly traded companies or publicly documented products. No personally identifiable information was collected or analyzed.

\section*{Conflict of Interest Disclosure}

The authors declare no financial conflicts of interest. Zhizibianjie is used as a case study example to illustrate theoretical constructs and does not represent a commercial relationship with the research team. The selection of Zhizibianjie was based on its empirical characteristics (extreme manifestation of Cultural Encoding) rather than commercial considerations.

\bibliographystyle{apalike}
\bibliography{references}

\end{document}